\documentclass[11pt,a4paper]{article}
\usepackage[hyperref]{acl2018}
\usepackage{times}
\usepackage{latexsym}
\usepackage{amsmath, bm, amsfonts, wrapfig, graphicx, multirow}

\usepackage{url}

\aclfinalcopy 
\def\aclpaperid{1181} 

\def\figref#1{Fig.~\ref{#1}}
\def\secref#1{Sec.~\ref{#1}}
\def\tabref#1{Table~\ref{#1}}
\def\eqnref#1{Eqn.~\ref{#1}}

\title{Fast Abstractive Summarization with \\ Reinforce-Selected Sentence Rewriting}

\author{Yen-Chun Chen \and Mohit Bansal \\
  UNC Chapel Hill \\
  {\tt \{yenchun, mbansal\}@cs.unc.edu} \\
 }

\date{}

\begin{document}
\maketitle

\begin{abstract}
Inspired by how humans summarize long documents, we propose an accurate and fast summarization model that first selects salient sentences and then rewrites them abstractively (i.e., compresses and paraphrases) to generate a concise overall summary.
We use a novel sentence-level policy gradient method to bridge the non-differentiable computation between these two neural networks in a hierarchical way, while maintaining language fluency.
Empirically, we achieve the new state-of-the-art on all metrics (including human evaluation) on the CNN/Daily Mail dataset, as well as significantly higher abstractiveness scores.
Moreover, by first operating at the sentence-level and then the word-level, 
we enable \emph{parallel decoding} of our neural generative model that results in substantially faster (10-20x) inference speed as well as 4x faster training convergence than previous long-paragraph encoder-decoder models.
We also demonstrate the generalization of our model on the test-only DUC-2002 dataset, where we achieve higher scores than a state-of-the-art model.

\end{abstract}

\section{Introduction}
The task of document summarization has two main paradigms: extractive and abstractive. 
The former method directly chooses and outputs the salient sentences (or phrases) in the original document~\citep{Jing:2000:CPB:974305.974329,Knight:2000:SSS:647288.721086,Martins:2009:SJM:1611638.1611639,Berg-Kirkpatrick:2011:JLE:2002472.2002534}.
The latter abstractive approach involves rewriting the summary~\citep{Banko:2000:HGB:1075218.1075259,topiary}, and has seen substantial recent gains due to neural sequence-to-sequence models
~\citep{chopra-auli-rush:2016:N16-1,nallapati2016abstractive,get_to_the_point,DBLP:journals/corr/PaulusXS17}.
Abstractive models can be more concise by performing generation from scratch, but they suffer from slow and inaccurate encoding of very long documents, with the attention model being required to look at all encoded words (in long paragraphs) for decoding each generated summary word (slow, one by one sequentially). Abstractive models also suffer from redundancy (repetitions), especially when generating multi-sentence summary.

To address both these issues and combine the advantages of both paradigms, we propose a hybrid extractive-abstractive architecture, with policy-based reinforcement learning (RL) to bridge together the two networks.
Similar to how humans summarize long documents, our model first uses an \emph{extractor agent} to select salient sentences or highlights, and then employs an \emph{abstractor network} to rewrite (i.e., compress and paraphrase) each of these extracted sentences.
To overcome the non-differentiable behavior of our extractor and train on available document-summary pairs without saliency label, we next use actor-critic policy gradient with sentence-level metric rewards to connect these two neural networks and to learn sentence saliency.
We also avoid common language fluency issues~\citep{DBLP:journals/corr/PaulusXS17} by preventing the policy gradients from affecting the abstractive summarizer's word-level training, which is supported by our human evaluation study.
Our sentence-level reinforcement learning takes into account the word-sentence hierarchy, which better models the language structure and makes parallelization possible. Our extractor combines reinforcement learning and pointer networks, which is inspired by~\citet{RL_comb:45821}'s attempt to solve the Traveling Salesman Problem. Our abstractor is a simple encoder-aligner-decoder model (with copying) and is trained on pseudo document-summary sentence pairs obtained via simple automatic matching criteria.

Thus, our method incorporates the abstractive paradigm's advantages of concisely rewriting sentences and generating novel words from the full vocabulary, yet it adopts intermediate extractive behavior to improve the overall model's quality, speed, and stability. Instead of encoding and attending to every word in the long input document sequentially, our model adopts a human-inspired \emph{coarse-to-fine} approach that first extracts all the salient sentences and then decodes (rewrites) them (\emph{in parallel}).
This also avoids almost all redundancy issues because the model has already chosen non-redundant salient sentences to abstractively summarize (but adding an optional final reranker component does give additional gains by removing the fewer across-sentence repetitions).

Empirically, our approach is the new state-of-the-art on all ROUGE metrics \citep{lin:2004:ACLsummarization} as well as on METEOR \cite{denkowski:lavie:meteor-wmt:2014} of the CNN/Daily Mail dataset, achieving statistically significant improvements over previous models that use complex long-encoder, copy, and coverage mechanisms~\citep{get_to_the_point}.
The test-only DUC-2002 improvement also shows our model's better generalization than this strong abstractive system.
In addition, we surpass the popular lead-3 baseline on all ROUGE scores with an abstractive model. Moreover, our sentence-level abstractive rewriting module also produces substantially more (3x) novel $N$-grams that are not seen in the input document, as compared to the strong flat-structured model of~\newcite{get_to_the_point}.
This empirically justifies that our RL-guided extractor has learned sentence saliency, rather than benefiting from simply copying longer sentences.
We also show that our model maintains the same level of fluency as a conventional RNN-based model because the reward does not leak to our abstractor's word-level training.
Finally, our model's training is 4x and inference is more than 20x faster than the previous state-of-the-art. The optional final reranker gives further improvements while maintaining a 7x speedup.

Overall, our contribution is three fold: 
First we propose a novel sentence-level RL technique for the well-known task of abstractive summarization, effectively utilizing the word-then-sentence hierarchical structure without annotated matching sentence-pairs between the document and ground truth summary.
Next, our model achieves the new state-of-the-art on all metrics of multiple versions of a popular summarization dataset (as well as a test-only dataset) both extractively and abstractively, without loss in language fluency (also demonstrated via human evaluation and abstractiveness scores). 
Finally, our parallel decoding results in a significant 10-20x speed-up over the previous best neural abstractive summarization system with even better accuracy.\footnote{We are releasing our code, best pretrained models, as well as output summaries, to promote future research: \url{https://github.com/ChenRocks/fast_abs_rl}}

\section{Model}
In this work, we consider the task of summarizing a given long text document
into several (ordered) highlights, which are then combined to form a multi-sentence summary.
Formally, given a training set of document-summary pairs $\{x_i, y_i\}_{i=1}^{N}$, 
our goal is to approximate the function
$h: X \rightarrow Y, X = \{x_i\}_{i=1}^N, Y = \{y_i\}_{i=1}^N$
such that $ h(x_i) = y_i, 1 \le i \le N $.
Furthermore, we assume there exists an abstracting function $g$ defined as:
$ \forall s \in S_i, \exists d \in D_i $
such that $ g(d) = s , 1 \le i \le N$,
where $S_i$ is the set of summary sentences in $x_i$ and
$D_i$ the set of document sentences in $y_i$.
i.e., in any given pair of document and summary, every summary sentence can be produced from some document sentence.
For simplicity, we omit subscript $i$ in the remainder of the paper.
Under this assumption, we can further define another latent function
$f: X \rightarrow D^n$ that satisfies
$ 
f(x) = \{d_t\}_{j=1}^n
\text{ and } 
y = h(x) = [g(d_1), g(d_2), \dots, g(d_n)]
$,
where $[,]$ denotes sentence concatenation.
This latent function $f$ can be seen as an extractor that chooses the salient (ordered) sentences in a given document for the abstracting function $g$ to rewrite.
Our overall model consists of these two submodules,
the \textit{extractor agent} and the \textit{abstractor network},
to approximate the above-mentioned $f$ and $g$, respectively.

\begin{figure*}
 \centering
 \includegraphics[width=0.9\textwidth]{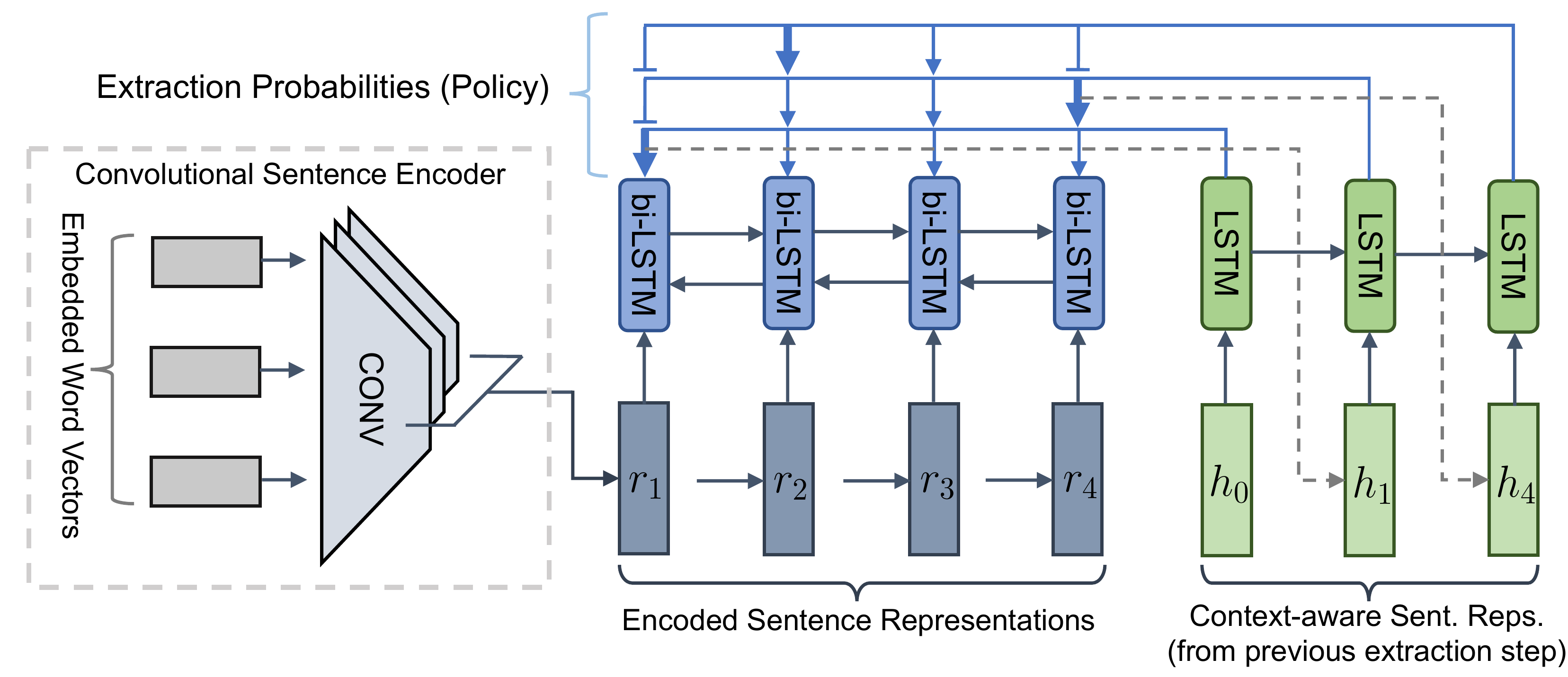}
 \vspace{-0.3cm}
 \caption{Our extractor agent: the convolutional encoder computes representation $r_j$ for each sentence. The RNN encoder (blue) computes context-aware representation $h_j$ and then the RNN decoder (green) selects sentence $j_t$ at time step $t$. With $j_t$ selected, $h_{j_t}$ will be fed into the decoder at time $t+1$. 
 }
 \vspace{-0.2cm}
 \label{fig:model}
\end{figure*}

\subsection{Extractor Agent}
The extractor agent is designed to model $f$, which can be thought of as extracting salient sentences from the document.
We exploit a hierarchical neural model to learn the sentence representations of the document and a `selection network' to extract sentences based on their representations.

\subsubsection{Hierarchical Sentence Representation}
\label{sec:sent-rep}
We use a temporal convolutional model~\cite{kim:2014:EMNLP2014} to compute $r_j$, the representation of 
each individual sentence in the documents (details in supplementary). 
To further incorporate global context of the document and capture the long-range semantic dependency between sentences, 
a bidirectional LSTM-RNN \citep{Hochreiter:1997:LSM:265493.264179,Schuster97bidirectionalrecurrent}
is applied on the convolutional output.
This enables learning a strong representation,
denoted as $h_j$ for the $j$-th sentence in the document, that takes into account the context of all previous and future sentences in the same document.

\subsubsection{Sentence Selection}\label{subsec:sent_select}

Next, to select the extracted sentences based on the above sentence representations, 
we add another LSTM-RNN to train a \textit{Pointer Network}~\citep{NIPS2015_5866:pointer_networks},
to extract sentences recurrently.
We calculate the extraction probability by:
\vspace{-5pt}
\begin{equation} 
\label{eq:ptrnet}
u^t_j = \begin{cases}
  v_p^\top \tanh(
  W_{p1}h_j + W_{p2}e_t) & 
        \text{if } j_t \ne j_k \\
        & \quad \forall k < t\\
  -\infty & \text{otherwise}
\end{cases} \\
\vspace{-5pt}
\end{equation}
\vspace{-5pt}
\begin{equation}
\label{eq:extraction_prob}
P(j_t | j_1, \dots, j_{t-1}) = \text{softmax}(u^t)
\vspace{-5pt}
\end{equation}
where $e_t$'s are the output of the \textit{glimpse} operation~\cite{OrderMatters}:
\vspace{-3pt}
\begin{align} \label{eq:glimpse}
  a^t_j &=
    v_g^\top \tanh(W_{g1}h_j + W_{g2}z_t) \\
  \alpha ^t &= \text{softmax}(a^t) \\
  e_t &= \sum_j \alpha^t_j W_{g1} h_j
  \vspace{-5pt}
\end{align}
In \eqnref{eq:glimpse}, $z_t$ is the output of the added LSTM-RNN (shown in green in \figref{fig:model}) which is 
referred to as the \textit{decoder}.
All the $W$'s and $v$'s are trainable parameters.
At each time step $t$, the decoder performs a 2-hop attention mechanism:
It first attends to $h_j$'s to get a context vector $e_t$ and then attends to $h_j$'s again for the extraction probabilities.\footnote{Note that we force-zero the extraction prob. of already extracted sentences so as to prevent the model from using repeating document sentences and suffering from redundancy. This is non-differentiable and hence only done in RL training.}
This model is essentially classifying all sentences of the document at each extraction step.
An illustration of the whole extractor is shown in \figref{fig:model}.

\begin{figure}
 \centering
 \includegraphics[width=0.5\textwidth]{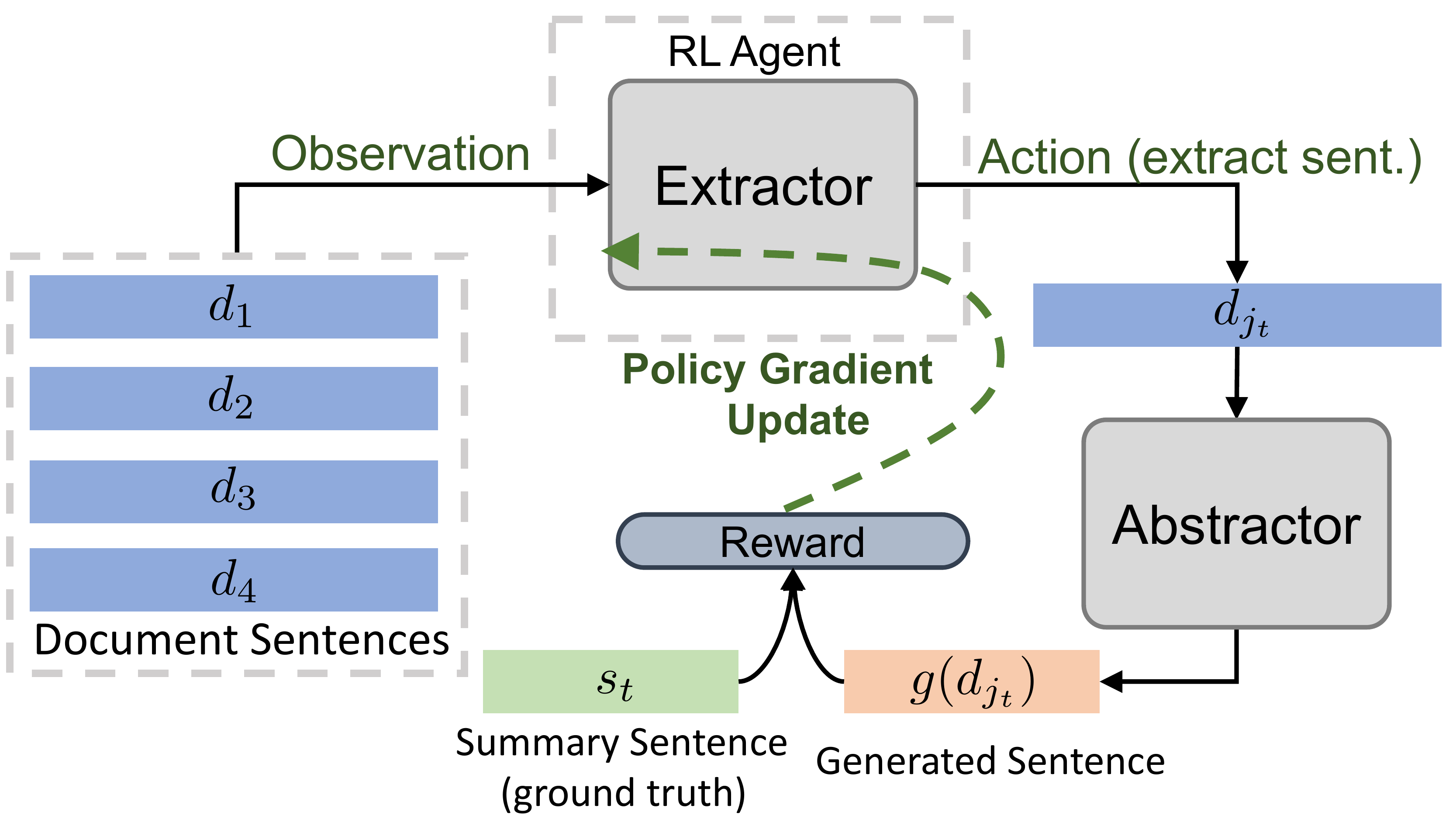}
 \vspace{-0.9cm}
 \caption{
 Reinforced training of the extractor (for one extraction step) and its interaction with the abstractor.
 For simplicity, the critic network is not shown.
 Note that all $d$'s and $s_t$ are raw sentences, \emph{not} vector representations.
 }
 \vspace{-0.2cm}
 \label{fig:pipeline}
\end{figure}

\subsection{Abstractor Network}
\label{sec:abs}
The abstractor network approximates $g$, 
which compresses and paraphrases an extracted document sentence to a concise summary sentence.
We use the standard encoder-aligner-decoder~\cite{bahdanau+al-2014-nmt,luong-pham-manning:2015:EMNLP}. We add the copy mechanism\footnote{We use the terminology of \textit{copy mechanism} (originally named 
\textit{pointer-generator}) in order to avoid confusion with the \textit{pointer network}~\cite{NIPS2015_5866:pointer_networks}.}
to help directly copy some out-of-vocabulary (OOV) words~\cite{get_to_the_point}.
For more details, please refer to the supplementary.

\section{Learning}
Given that our extractor performs a non-differentiable \emph{hard} extraction, 
we apply standard policy gradient methods to bridge the back-propagation and form an end-to-end trainable (stochastic) computation graph.
However, simply starting from a randomly initialized network to train the whole model in an end-to-end fashion is infeasible. 
When randomly initialized, the extractor would often select sentences that are not relevant,
so it would be difficult for the abstractor to learn to abstractively rewrite.
On the other hand, without a well-trained abstractor the extractor would get noisy reward, which leads to a bad estimate of the policy gradient and a sub-optimal policy.
We hence propose optimizing each sub-module separately using maximum-likelihood (ML) objectives: 
train the extractor to select salient sentences (fit $f$) and the abstractor to generate shortened summary (fit $g$).
Finally, RL is applied to train the full model end-to-end (fit $h$).

\subsection{Maximum-Likelihood Training for Submodules}

\noindent\textbf{Extractor Training}: In ~\secref{subsec:sent_select}, we have formulated our sentence selection as classification. 
However, most of the summarization datasets are end-to-end document-summary pairs without extraction (saliency) labels for each sentence.
Hence, we propose a simple similarity method to provide a `proxy' target label for the extractor. Similar to the extractive model of~\newcite{AAAI17:summarunner}, for each ground-truth summary sentence, we find the most similar document sentence $d_{j_t}$ by:\footnote{\newcite{AAAI17:summarunner} 
selected sentences greedily to maximize the global summary-level ROUGE, whereas we match exactly 1 document sentence for each GT summary sentence based on the \textit{individual} sentence-level score.}
\vspace{-5pt}
\begin{equation} 
\label{eq:proxy-extract}
j_t = \text{argmax}_i(\text{ROUGE-L}_{recall}(d_i, s_t))
\vspace{-5pt}
\end{equation}
Given these proxy training labels, the extractor is then trained to minimize the cross-entropy loss.

\noindent\textbf{Abstractor Training}: For the abstractor training, we create training pairs by taking each summary sentence and pairing it with its extracted document sentence (based on \eqnref{eq:proxy-extract}). The network is trained as an usual sequence-to-sequence model to minimize 
the cross-entropy loss
$
L(\theta_{abs}) = - \frac{1}{M}\sum_{m=1}^{M}\text{log}P_{\theta_{abs}}(w_m|w_{1:m-1})
$
of the decoder language model at each generation step,
where $\theta_{abs}$ is the set of trainable parameters of the abstractor and $w_m$ the $m^{th}$ generated word.

\subsection{Reinforce-Guided Extraction}
\label{sec:rl-ext}
Here we explain how policy gradient techniques are applied to optimize the whole model.
To make the extractor an RL agent, we can formulate a Markov Decision Process (MDP)\footnote{Strictly speaking, this is a \textit{Partially Observable Markov Decision Process} (POMDP). 
We approximate it as an MDP by assuming that the RNN hidden state contains all past info.}: 
at each extraction step $t$, the agent observes the current state $c_t = (D, d_{j_{t-1}})$, 
samples an action 
$j_t \sim \pi_{\theta_a, \omega}(c_t, j)
= P(j)$ from \eqnref{eq:extraction_prob}
to extract a document sentence and 
receive a reward\footnote{In \eqnref{eq:proxy-extract}, we use ROUGE-recall because we want the extracted sentence to contain as much information as possible for rewriting.
Nevertheless, for~\eqnref{eq:step-reward}, ROUGE-$F_1$ is more suitable because the abstractor $g$ is supposed to rewrite the extracted sentence $d$ to be as \textit{concise} as the ground truth $s$.
}
\vspace{-5pt}
\begin{equation}
\label{eq:step-reward}
r(t+1) = \text{ROUGE-L}_{F_1}(g(d_{j_t}), s_t)
\vspace{-5pt}
\end{equation}
after the abstractor summarizes the extracted sentence $d_{j_t}$.
We denote the trainable parameters of the extractor agent 
by $\theta = \{\theta_a, \omega \}$ for the decoder and hierarchical encoder respectively.
We can then train the extractor with policy-based RL.
We illustrate this process in~\figref{fig:pipeline}.

The vanilla policy gradient algorithm, REINFORCE~\citep{Williams:1992:SSG:139611.139614}, is known for high variance.
To mitigate this problem, we add a critic network with trainable parameters $\theta_c$ to predict the state-value function $V^{\pi_{\theta_a, \omega}}(c)$.
The predicted value of critic $b_{\theta_c, \omega}(c)$ is called the `baseline',
which is then used to estimate the \textit{advantage function}:
$
A^{\pi_\theta}(c, j) = 
Q^{\pi_{\theta_a, \omega}}(c, j) - 
V^{\pi_{\theta_a, \omega}}(c)
$
because the total return $R_t$ is an estimate of action-value function $Q(c_t, j_t)$.
Instead of maximizing $Q(c_t, j_t)$ as done in REINFORCE, we maximize $A^{\pi_\theta}(c, j)$ with the following policy gradient:
\vspace{-5pt}
\begin{equation}
\begin{split}
\nabla_{\theta_a, \omega} J(\theta_a, \omega) = \\
\mathbb{E}
[\nabla_{\theta_a, \omega}\text{log}\pi_\theta(c, j) A^{\pi_{\theta}}(c, j)]
\end{split}
\end{equation}
And the critic is trained to minimize the square loss:
$
L_c(\theta_c, \omega) = (b_{\theta_c, \omega}(c_t) - R_t)^2
$.
This is known as the \textit{Advantage Actor-Critic} (A2C), a synchronous variant of A3C \citep{pmlr-v48-mniha16}.
For more A2C details, please refer to the supp.

Intuitively, our RL training works as follow:
If the extractor chooses a good sentence, after the abstractor rewrites it the ROUGE match would be high and thus the action is encouraged.
If a bad sentence is chosen, though the abstractor still produces a compressed version of it, the summary would not match the ground truth and the low ROUGE score discourages this action.
Our RL with a sentence-level agent is a novel attempt in neural summarization.
We use RL as a saliency guide without altering the abstractor's language model, while previous work applied RL on the word-level, which could be prone to gaming the metric at the cost of language fluency.\footnote{During this RL training of the extractor, we keep the abstractor parameters fixed. 
Because the input sentences for the abstractor are extracted by an intermediate stochastic policy of the extractor, it is impossible to find the correct target summary for the abstractor to fit $g$ with ML objective. 
Though it is possible to optimize the abstractor with RL, 
in out preliminary experiments we found that this does not improve the overall ROUGE, most likely because this RL optimizes at a sentence-level and can add across-sentence redundancy. We achieve SotA results without this abstractor-level RL.}

\paragraph{Learning how many sentences to extract:}
\label{sec:end-of-extract}
In a typical RL setting like game playing, an episode is usually terminated by the environment.
On the other hand, in text summarization, the agent does not know in advance how many summary sentence to produce for a given article (since the desired length varies for different downstream applications).
We make an important yet simple, intuitive adaptation to solve this: by adding a `stop' action to the policy action space.
In the RL training phase, we add another set of trainable parameters $v_{EOE}$ (EOE stands for `End-Of-Extraction') with the same dimension as the sentence representation.
The pointer-network decoder treats $v_{EOE}$ as one of the extraction candidates and hence naturally results in a stop action in the stochastic policy.
We set the reward for the agent performing EOE to 
$\text{ROUGE-1}_{F_1}([\{g(d_{j_t})\}_t], [\{s_t\}_t])$; whereas for any extraneous, unwanted extraction step, the agent receives zero reward.
The model is therefore encouraged to extract when there are still remaining ground-truth summary sentences (to accumulate intermediate reward),
and learn to stop by optimizing a global ROUGE and avoiding extra extraction.\footnote{We use ROUGE-1 for terminal reward because it is a better measure of bag-of-words information
(i.e., has all the important information been generated);
while ROUGE-L is used as intermediate rewards since it is known for better measurement of language fluency within a local sentence.}
Overall, this modification allows dynamic decisions of number-of-sentences based on the input document, eliminates the need for tuning a fixed number of steps, and enables a data-driven adaptation for any specific dataset/application.

\subsection{Repetition-Avoiding Reranking}
\label{sec:rerank}
Existing abstractive summarization systems on long documents suffer from generating repeating and redundant words and phrases.
To mitigate this issue, \citet{get_to_the_point} propose the coverage mechanism and \citet{DBLP:journals/corr/PaulusXS17} incorporate tri-gram avoidance during beam-search at test-time.
Our model without these already performs well because the summary sentences are generated from mutually exclusive document sentences, which naturally avoids redundancy.
However, we do get a small further boost to the summary quality by removing a few `across-sentence' repetitions, via a simple reranking strategy: At sentence-level, we apply the same beam-search tri-gram avoidance \citep{DBLP:journals/corr/PaulusXS17}.
We keep all $k$ sentence candidates generated by beam search, where $k$ is the size of the beam.
Next, we then rerank all $k^n$ combinations of the $n$ generated summary sentence beams.
The summaries are reranked by the number of repeated $N$-grams, the smaller the better.
We also apply the diverse decoding algorithm described in~\newcite{diverse} (which has almost no computation overhead) so as to get the above approach to produce useful diverse reranking lists. 
We show how much the redundancy affects the summarization task in \secref{sec:rerank-exp}.

\section{Related Work}
Early summarization works mostly focused on extractive and compression based methods
\citep{Jing:2000:CPB:974305.974329,Knight:2000:SSS:647288.721086,Clarke:Lapata:2010,Berg-Kirkpatrick:2011:JLE:2002472.2002534,filippova:43852}.
Recent large-sized corpora attracted neural methods for abstractive summarization
\citep{rush-chopra-weston:2015:EMNLP,chopra-auli-rush:2016:N16-1}.
Some of the recent success in neural abstractive models include hierarchical attention~\cite{nallapati2016abstractive},
coverage~\cite{Suzuki2016Summ,Chen2016DistractionBasedNN,get_to_the_point}, RL based metric optimization~\cite{DBLP:journals/corr/PaulusXS17}, graph-based attention~\cite{graph_attn_Tan2017AbstractiveDS}, and the copy mechanism~ \cite{DBLP:journals/corr/MiaoB16,gu-EtAl:2016:P16-1:copying_mechanism,get_to_the_point}.

Our model shares some high-level intuition with extract-then-compress methods.
Earlier attempts in this paradigm used Hidden Markov Models and rule-based systems \citep{Jing:2000:CPB:974305.974329},
statistical models based on parse trees~\citep{Knight:2000:SSS:647288.721086},
and integer linear programming based methods
\citep{Martins:2009:SJM:1611638.1611639,Gillick:2009:SGM:1611638.1611640,Clarke:Lapata:2010,Berg-Kirkpatrick:2011:JLE:2002472.2002534}.
Recent approaches investigated discourse structures
\citep{Louis:2010:DIC:1944506.1944533,hirao-EtAl:2013:EMNLP,kikuchi2014single,Wang:2015:SBT:2876444.2876454}, 
graph cuts~\citep{qian-liu:2013:EMNLP2}, 
and parse trees~\citep{D14-1076,conf/acl/BingLLLGP15}.
For neural models,~\citet{cheng-lapata:2016:P16-1} used a second neural net to select words from an extractor's output.
Our abstractor does not merely `compress' the sentences
but generatively produce novel words. 
Moreover, our RL bridges the extractor and the abstractor for end-to-end training.

Reinforcement learning has been used to optimize
the non-differential metrics of language generation and to mitigate exposure bias 
\citep{DBLP:journals/corr/RanzatoCAZ15,DBLP:journals/corr/BahdanauBXGLPCB16}. \citet{RL_TUD-CS-2015-0145} use Q-learning based RL for extractive summarization. \citet{DBLP:journals/corr/PaulusXS17} use RL policy gradient methods for abstractive summarization, utilizing sequence-level metric rewards with curriculum learning \citep{DBLP:journals/corr/RanzatoCAZ15} or weighted ML+RL mixed loss \citep{DBLP:journals/corr/PaulusXS17} for stability and language fluency.
We use sentence-level rewards to optimize the extractor while keeping our ML trained abstractor decoder fixed, so as to achieve the best of both worlds.

Training a neural network to use another fixed network has been investigated in machine translation
for better decoding \citep{DBLP:journals/corr/GuCL17} and real-time translation \citep{DBLP:journals/corr/GuNCL16}. 
They used a fixed pretrained \textit{translator} and applied policy gradient techniques to train another task-specific network.
In question answering (QA),~\newcite{c2r:P17-1020} extract one sentence and then generate the answer from the sentence's vector representation with RL bridging.
Another recent work attempted a new coarse-to-fine attention approach on summarization~\citep{C2F_Summ} and found desired sharp focus properties for scaling to larger inputs (though without metric improvements).
Very recently (concurrently), \citet{extract_rl} use RL for ranking sentences in pure extraction-based summarization and \citet{comm_agent} investigate multiple communicating encoder agents to enhance the copying abstractive summarizer.

Finally, there are some loosely-related recent works:
\citet{select-enc} proposed \textit{selective gate} to improve the attention in abstractive summarization.
\citet{s-net} used an \textit{extract-then-synthesis} approach on QA, 
where an extraction model predicts the important spans in the passage and then another synthesis model generates the final answer. 
\citet{cascade} attempted cascaded non-recurrent small networks on extractive QA, resulting a scalable, parallelizable model.
\citet{control-summ} added controlling parameters to adapt the summary to length, style, and entity preferences.
However, none of these used RL to bridge the non-differentiability of neural models.

\begin{table*}[t]
\begin{small}
\centering
\begin{tabular}{ | l | c  c  c | c |}
  \hline
  Models & \ \  ROUGE-1 \ \  &  \ \ ROUGE-2 \ \  & \ \  ROUGE-L \ \  &  \ \ METEOR \ \   \\
  \hline
  \multicolumn{5}{|c|} {Extractive Results} \\
  \hline
  lead-3~\citep{get_to_the_point} & 40.34 & 17.70 & 36.57 & 22.21 \\
  \citet{extract_rl} (sentence ranking RL) & 40.0 & 18.2 & 36.6 & - \\
  ff-ext & 40.63 & 18.35 & 36.82 & \textbf{22.91} \\
  rnn-ext  & 40.17 & 18.11 & 36.41 & 22.81 \\
  rnn-ext + RL &  \textbf{41.47} & \textbf{18.72} &  \textbf{37.76} & 22.35 \\
  \hline
  \multicolumn{5}{|c|} {Abstractive Results} \\
  \hline
  \citet{get_to_the_point} (w/o coverage) \ \ \ \ \ \ \ \ & 36.44 & 15.66 & 33.42 & 16.65\\ 
  \citet{get_to_the_point} & 39.53 & 17.28 & 36.38 & 18.72\\
  \citet{control-summ} (controlled) & 39.75 & 17.29 & 36.54 & - \\
  ff-ext + abs & 39.30 & 17.02 & 36.93 & 20.05 \\
  rnn-ext + abs & 38.38 & 16.12 & 36.04 & 19.39 \\
  rnn-ext + abs + RL &  {40.04} & {17.61} &  {37.59} & \textbf{21.00} \\
  rnn-ext + abs + RL + rerank \ \ \ \  &  \textbf{40.88} & \textbf{17.80} &  \textbf{38.54} & {20.38} \\ \hline
\end{tabular}
\vspace{-6pt}
\caption{
Results on the original, non-anonymized CNN/Daily Mail dataset.
Adding RL gives statistically significant improvements for all metrics over non-RL rnn-ext models (and over the state-of-the-art \citet{get_to_the_point}) in both extractive and abstractive settings with $p < 0.01$.
Adding the extra reranking stage yields statistically significant better results in terms of all ROUGE metrics with $p < 0.01$.
}
\vspace{-5pt}
\label{tab:raw}
\end{small}
\end{table*}

\section{Experimental Setup}

Please refer to the supplementary for full training details (all hyperparameter tuning was performed on the validation set).
We use the CNN/Daily Mail dataset~\cite{nips15_hermann} modified for summarization~\cite{nallapati2016abstractive}.
Because there are two versions of the dataset, original text and entity anonymized,
we show results on both versions of the dataset for a fair comparison to prior works. 
The experiment runs training and evaluation for each version separately.
Despite the fact that the 2 versions have been considered separately by the summarization community as 2 different datasets, we use same hyper-parameter values for both dataset versions to show the generalization of our model.
We also show improvements on the DUC-2002 dataset in a \emph{test-only} setup.

\subsection{Evaluation Metrics}
For all the datasets, we evaluate standard ROUGE-1, ROUGE-2, and ROUGE-L \citep{lin:2004:ACLsummarization} on full-length $F_1$ (with stemming) following previous works 
\citep{AAAI17:summarunner,get_to_the_point,DBLP:journals/corr/PaulusXS17}. 
Following \citet{get_to_the_point}, we also evaluate on METEOR \citep{denkowski:lavie:meteor-wmt:2014} for a more thorough analysis.

\subsection{Modular Extractive vs. Abstractive}
Our hybrid approach is capable of both extractive and abstractive (i.e., rewriting every sentence) summarization.
The extractor alone performs extractive summarization. 
To investigate the effect of the recurrent extractor (rnn-ext), we implement a feed-forward extractive baseline ff-ext (details in supplementary). 
It is also possible to apply RL to extractor without using the abstractor (rnn-ext + RL).\footnote{In this case the abstractor function $g(d) = d$.}
Benefiting from the high modularity of our model, we can make our summarization system abstractive by 
simply applying the abstractor on the extracted sentences.
Our abstractor rewrites each sentence and generates novel words from a large vocabulary, and hence every word in our overall summary is generated from scratch;
making our full model categorized into the abstractive paradigm.\footnote{Note that the abstractive CNN/DM dataset does \emph{not} include any human-annotated extraction label, and hence our models do not receive any direct extractive supervision.}
We run experiments on separately trained extractor/abstractor (ff-ext + abs, rnn-ext + abs) and the reinforced full model (rnn-ext + abs + RL) as well as the final reranking version (rnn-ext + abs + RL + rerank).

\section{Results}

For easier comparison, we show separate tables for the original-text vs. anonymized versions -- \tabref{tab:raw} and \tabref{tab:anon}, respectively.
Overall, our model achieves strong improvements and the new state-of-the-art on both extractive and abstractive settings for both versions of the CNN/DM dataset (with some comparable results on the anonymized version).
Moreover, \tabref{tab:duc} shows the generalization of our abstractive system to an out-of-domain test-only setup (DUC-2002), where our model achieves better scores than \citet{get_to_the_point}.

\begin{table}[t]
\begin{small}
\centering
\begin{tabular}{ | l | c  c  c|}
  \hline
  Models & R-1 & R-2 & R-L \\
  \hline
  \multicolumn{4}{|c|} {Extractive Results} \\
  \hline
  lead-3 \citep{AAAI17:summarunner} & 39.2 & 15.7 & 35.5 \\
  \citet{AAAI17:summarunner} & 39.6 & 16.2 & 35.3 \\
  ff-ext & 39.51 & \textbf{16.85} & 35.80 \\
  rnn-ext  & 38.97 & 16.65 & 35.32 \\
  rnn-ext + RL & \textbf{40.13} & 16.58 & \textbf{36.47} \\
  \hline
  \multicolumn{4}{|c|} {Abstractive Results} \\
  \hline
  \citet{nallapati2016abstractive}
  & 35.46 & 13.30 & 32.65 \\
  \citet{control-summ} (controlled)
  & 38.68 & 15.40 & 35.47 \\
  \citet{DBLP:journals/corr/PaulusXS17} (ML)
  & 38.30 & 14.81 & 35.49 \\
  \citet{DBLP:journals/corr/PaulusXS17} (RL+ML) & \textbf{39.87} & 15.82 & 36.90 \\
  ff-ext + abs & 38.73 & 15.70 & 36.33 \\
  rnn-ext + abs & 37.58 & 14.68 & 35.24 \\
  rnn-ext + abs + RL  & 38.80 & 15.66 & 36.37 \\
  rnn-ext + abs + RL + rerank & 39.66 & \textbf{15.85} & \textbf{37.34} \\ \hline
\end{tabular}
\vspace{-5pt}
\caption{ROUGE for anonymized CNN/DM.}
\vspace{-8pt}
\label{tab:anon}
\end{small}
\end{table}

\subsection{Extractive Summarization}
In the extractive paradigm, 
we compare our model with 
the extractive model from \citet{AAAI17:summarunner} and a strong lead-3 baseline.
For producing our summary, we simply concatenate the extracted sentences from the extractors.
From~\tabref{tab:raw} and ~\tabref{tab:anon},
we can see that our feed-forward extractor out-performs the lead-3 baseline, 
empirically showing that our hierarchical sentence encoding model is capable of extracting salient sentences.\footnote{The ff-ext model outperforms rnn-ext possibly because it does not predict sentence ordering; thus is easier to optimize and the n-gram based metrics do not consider sentence ordering. Also note that in our MDP formulation, we cannot apply RL on ff-ext due to its historyless nature.
Even if applied naively, there is no mean for the feed-forward model to learn the EOE described in~\secref{sec:end-of-extract}.}
The reinforced extractor performs the best, because of the ability to get the summary-level reward and the reduced train-test mismatch of feeding the previous extraction decision.
The improvement over lead-3 is consistent across both tables.
In~\tabref{tab:anon}, it outperforms the previous best neural extractive model~\cite{AAAI17:summarunner}.
In~\tabref{tab:raw}, our model also outperforms a recent, concurrent sentence-ranking RL model by~\citet{extract_rl}, showing that our pointer-network extractor and reward formulations are very effective when combined with A2C RL.

\subsection{Abstractive Summarization}
After applying the abstractor, the ff-ext based model still out-performs the rnn-ext model.
Both combined models exceed the pointer-generator model \citep{get_to_the_point} without coverage by a large margin for all metrics, 
showing the effectiveness of our 2-step hierarchical approach:
our method naturally avoids repetition by extracting multiple sentences with different keypoints.\footnote{A trivial \textit{lead-3 + abs} baseline obtains ROUGE of (37.37, 15.59, 34.82), which again confirms the importance of our reinforce-based sentence selection.}

Moreover, after applying reinforcement learning, our model performs better than the best model of \citet{get_to_the_point} and the best ML trained model of \citet{DBLP:journals/corr/PaulusXS17}.
Our reinforced model outperforms the ML trained rnn-ext + abs baseline with statistical significance of $p < 0.01$ on all metrics for both version of the dataset, indicating the effectiveness of the RL training.
Also, rnn-ext + abs + RL is statistically significant better than \citet{get_to_the_point} for all metrics with $p < 0.01$.\footnote{We calculate statistical significance based on the bootstrap test~\cite{noreen1989computer,efron1994introduction} with 100K samples. Output of \citet{DBLP:journals/corr/PaulusXS17} is not available so we couldn't test for statistical significance there.}
In the supplementary, we show the learning curve of our RL training, where the average reward goes up quickly after the extractor learns the End-of-Extract action and then stabilizes.
For all the above models, we use standard greedy decoding and find that it performs well.

\begin{table}[t]
\begin{small}
\centering
\begin{tabular}{ | l | c  c  c |}
  \hline
  Models & R-1 & R-2 & R-L \\
  \hline
  \citet{get_to_the_point} & 37.22 & 15.78 & 33.90 \\
  rnn-ext + abs + RL  & 39.46 & 17.34 & 36.72 \\ \hline
\end{tabular}
\vspace{-8pt}
\caption{Generalization to DUC-2002 (F1). }
\vspace{-12pt}
\label{tab:duc}
\end{small}
\end{table}

\paragraph{Reranking and Redundancy}
\label{sec:rerank-exp}
Although the extract-then-abstract approach inherently will not generate repeating sentences like other neural-decoders do, there might still be across-sentence redundancy because the abstractor is not aware of other extracted sentences when decoding one.
Hence, we incorporate an optional reranking strategy described in~\secref{sec:rerank}.
The improved ROUGE scores indicate that this successfully removes some remaining redundancies and hence produces more concise summaries.
Our best abstractive model (rnn-ext + abs + RL + rerank) is clearly superior than the one of \citet{get_to_the_point}.
We are comparable on R-1 and R-2 but a 0.4 point improvement on R-L w.r.t. \citet{DBLP:journals/corr/PaulusXS17}.\footnote{We do not list the scores of their pure RL model because they discussed its bad readability.} We also outperform the results of~\citet{control-summ} on both original and anonymized dataset versions.
Several previous works have pointed out that extractive baselines are very difficult to beat (in terms of ROUGE) by an abstractive system \citep{get_to_the_point,AAAI17:summarunner}. Note that our best model is one of the first abstractive models to outperform the lead-3 baseline on the original-text CNN/DM dataset. 
Our extractive experiment serves as a complementary analysis of the effect of RL with extractive systems.

\begin{table}[t]
\centering
\resizebox{\columnwidth}{!}{
\begin{tabular}{ | l || c   c | c|}
  \hline
   & Relevance & Readability & Total \\ \hline
  \citet{get_to_the_point} & 120 & 128 & 248 \\
  rnn-ext + abs + RL + rerank & \textbf{137} & \textbf{133} & \textbf{270} \\ \hline
  Equally good/bad & 43 & 39 & 82 \\
  \hline
\end{tabular}}
\vspace{-12pt}
\caption{
Human Evaluation: pairwise comparison between our final model and~\newcite{get_to_the_point}.
}
\vspace{-10pt}
\label{tab:human_eval}
\end{table}

\subsection{Human Evaluation}
\label{sec:human_eval}
We also conduct human evaluation to ensure robustness of our training procedure.
We measure \emph{relevance} and \emph{readability} of the summaries.
Relevance is based on the summary containing important, salient information from the input article, being correct by avoiding contradictory/unrelated information, and avoiding repeated/redundant information. Readability is based on the summary’s fluency, grammaticality, and coherence.
To evaluate both these criteria, we design the following Amazon MTurk experiment: we randomly select 100 samples from the CNN/DM test set and ask the human testers (3 for each sample)
to rank between summaries (for relevance and readability) produced by our model and that of \citet{get_to_the_point}
(the models were anonymized and randomly shuffled), i.e. A is better, B is better, both are equally good/bad.
Following previous work, the input article and ground truth summaries are also shown to the human participants in addition to the two model summaries.\footnote{We selected human annotators that were located in the US, had
an approval rate greater than 95\%, and had at least 10,000 approved HITs on record.}
From the results shown in~\tabref{tab:human_eval}, we can see that our model is better in both relevance and readability w.r.t. \citet{get_to_the_point}.

\begin{table}[t]
\centering
\resizebox{\columnwidth}{!}{
\begin{tabular}{ | l || c | c |}
  \hline
  & \multicolumn{2}{c|} {Speed} \\ \hline
  Models & total time (hr) & words / sec \\ \hline
  \cite{get_to_the_point} & 12.9 & 14.8 \\
  rnn-ext + abs + RL & 0.68 & 361.3 \\
  rnn-ext + abs + RL + rerank & 2.00 (1.46 +0.54) & 109.8 \\ \hline
\end{tabular}}
\vspace{-12pt}
\caption{
Speed comparison with \protect\newcite{get_to_the_point}.
}
\vspace{-12pt}
\label{tab:speed}
\end{table}

\subsection{Speed Comparison}

Our two-stage extractive-abstractive hybrid model is not only the SotA on summary quality metrics, but more importantly also gives a significant speed-up in both train and test time over a strong neural abstractive system~\citep{get_to_the_point}.\footnote{The only publicly available code with a pretrained model for neural summarization which we can test the speed.}

Our full model is composed of a extremely fast extractor and a parallelizable abstractor, where the computation bottleneck is on the abstractor, which has to generate summaries with a large vocabulary from scratch.\footnote{The time needed for extractor is negligible w.r.t. the abstractor because it does not require large matrix multiplication for generating every word.
Moreover, with convolutional encoder at word-level made parallelizable by the hierarchical rnn-ext, our model is scalable for very long documents.}
The main advantage of our abstractor at decoding time is that we can first compute all the extracted sentences for the document, and then abstract every sentence concurrently (\emph{in parallel}) to generate the overall summary.
In \tabref{tab:speed}, we show the substantial test-time speed-up of our model compared to \citet{get_to_the_point}.\footnote{For details of training speed-up, please see the supp.}
We calculate the total decoding time for producing all summaries for the test set.\footnote{We time the model of \citet{get_to_the_point} using beam size of 4 (used for their best-reported scores).
Without beam-search, it gets significantly worse ROUGE of (36.62, 15.12, 34.08), so we do not compare speed-ups w.r.t. that version.}
Due to the fact that the main test-time speed bottleneck of RNN language generation model is that the model is constrained to generate one word at a time,
the total decoding time is dependent on the number of total words generated; we hence also report the decoded words per second for a fair comparison.
Our model without reranking is extremely fast. 
From \tabref{tab:speed} we can see that we achieve a speed up of 18x in time and 24x in word generation rate. 
Even after adding the (optional) reranker, we still maintain a 6-7x speed-up (and hence a user can choose to use the reranking component depending on their downstream application's speed requirements).\footnote{Most of the recent neural abstractive summarization systems are of similar algorithmic complexity to that of~\citet{get_to_the_point}.
The main differences such as the training objective (ML vs. RL) and copying (soft/hard) has negligible test runtime compared to the slowest component: the long-summary attentional-decoder's sequential generation;
and this is the component that we substantially speed up via our parallel sentence decoding with sentence-selection RL.}

\begin{table}[t]
\centering
\resizebox{\columnwidth}{!}{
\begin{tabular}{ | l || c  c  c  c|}
  \hline
  & \multicolumn{4}{c|}{Novel $N$-gram (\%)} \\
  \hline
  Models & 1-gm & 2-gm & 3-gm & 4-gm \\ \hline
  \citet{get_to_the_point} & 0.1 & 2.2 & 6.0 & 9.7 \\
  rnn-ext + abs + RL + rerank & 0.3 & 10.0 & 21.7 & 31.6 \\ \hline
  reference summaries & 10.8 & 47.5 & 68.2 & 78.2 \\
  \hline
\end{tabular}}
\vspace{-10pt}
\caption{
Abstractiveness: novel $n$-gram counts.
}
\label{tab:novelty}
\vspace{-10pt}
\end{table}

\section{Analysis}

\subsection{Abstractiveness}
We compute an abstractiveness score~\cite{get_to_the_point}
as the ratio of novel $n$-grams in the generated summary that are not present
in the input document. The results are shown in~\tabref{tab:novelty}: our model rewrites substantially more abstractive summaries than previous work.
A potential reason for this is that when trained with individual sentence-pairs, the abstractor learns to drop more document words so as to write individual summary sentences as concise as human-written ones; thus the improvement in multi-gram novelty.

\subsection{Qualitative Analysis on Output Examples}
\label{sec:qualitative}

We show examples of how our best model selects sentences and then rewrites them.
In the supplementary \figref{fig:sample1} and \figref{fig:sample2}, we can see how the abstractor rewrites the extracted sentences concisely while keeping the mentioned facts.
Adding the reranker makes the output more compact globally.
We observe that when rewriting longer text, the abstractor would have many facts to choose from (\figref{fig:sample2} sentence 2) and
this is where the reranker helps avoid redundancy across sentences.

\section{Conclusion}
We propose a novel sentence-level RL model for abstractive summarization, which makes the model aware of the word-sentence hierarchy.
Our model achieves the new state-of-the-art on both CNN/DM versions as well a better generalization on test-only DUC-2002, along with a significant speed-up in training and decoding.

\section*{Acknowledgments}
We thank the anonymous reviewers for their helpful comments. This work was supported by a
Google Faculty Research Award, a Bloomberg Data Science Research Grant, an IBM Faculty
Award, and NVidia GPU awards.

\bibliography{acl2018}

\begin{thebibliography}{62}
\expandafter\ifx\csname natexlab\endcsname\relax\def\natexlab#1{#1}\fi

\bibitem[{Bahdanau et~al.(2017)Bahdanau, Brakel, Xu, Goyal, Lowe, Pineau,
  Courville, and Bengio}]{DBLP:journals/corr/BahdanauBXGLPCB16}
Dzmitry Bahdanau, Philemon Brakel, Kelvin Xu, Anirudh Goyal, Ryan Lowe, Joelle
  Pineau, Aaron~C. Courville, and Yoshua Bengio. 2017.
\newblock \href {http://arxiv.org/abs/1607.07086} {An actor-critic algorithm
  for sequence prediction}.
\newblock In \emph{ICLR}.

\bibitem[{Bahdanau et~al.(2015)Bahdanau, Cho, and
  Bengio}]{bahdanau+al-2014-nmt}
Dzmitry Bahdanau, Kyunghyun Cho, and Yoshua Bengio. 2015.
\newblock Neural machine translation by jointly learning to align and
  translate.
\newblock In \emph{ICLR}.

\bibitem[{Banko et~al.(2000)Banko, Mittal, and
  Witbrock}]{Banko:2000:HGB:1075218.1075259}
Michele Banko, Vibhu~O. Mittal, and Michael~J. Witbrock. 2000.
\newblock \href {https://doi.org/10.3115/1075218.1075259} {Headline generation
  based on statistical translation}.
\newblock In \emph{Proceedings of the 38th Annual Meeting on Association for
  Computational Linguistics}, ACL '00, pages 318--325, Stroudsburg, PA, USA.
  Association for Computational Linguistics.

\bibitem[{Bello et~al.(2017)Bello, Pham, Le, Norouzi, and
  Bengio}]{RL_comb:45821}
Irwan Bello, Hieu Pham, Quoc~V. Le, Mohammad Norouzi, and Samy Bengio. 2017.
\newblock \href {https://openreview.net/pdf?id=Bk9mxlSFx} {Neural combinatorial
  optimization with reinforcement learning}.
\newblock \emph{arXiv preprint 1611.09940}.

\bibitem[{Berg-Kirkpatrick et~al.(2011)Berg-Kirkpatrick, Gillick, and
  Klein}]{Berg-Kirkpatrick:2011:JLE:2002472.2002534}
Taylor Berg-Kirkpatrick, Dan Gillick, and Dan Klein. 2011.
\newblock \href {http://dl.acm.org/citation.cfm?id=2002472.2002534} {Jointly
  learning to extract and compress}.
\newblock In \emph{Proceedings of the 49th Annual Meeting of the Association
  for Computational Linguistics: Human Language Technologies - Volume 1}, HLT
  '11, pages 481--490, Stroudsburg, PA, USA. Association for Computational
  Linguistics.

\bibitem[{Bing et~al.(2015)Bing, Li, Liao, Lam, Guo, and
  Passonneau}]{conf/acl/BingLLLGP15}
Lidong Bing, Piji Li, Yi~Liao, Wai Lam, Weiwei Guo, and Rebecca~J. Passonneau.
  2015.
\newblock Abstractive multi-document summarization via phrase selection and
  merging.
\newblock In \emph{ACL}.

\bibitem[{{\c{C}}elikyilmaz et~al.(2018){\c{C}}elikyilmaz, Bosselut, He, and
  Choi}]{comm_agent}
Asli {\c{C}}elikyilmaz, Antoine Bosselut, Xiaodong He, and Yejin Choi. 2018.
\newblock \href {http://arxiv.org/abs/1803.10357} {Deep communicating agents
  for abstractive summarization}.
\newblock \emph{NAACL-HLT}.

\bibitem[{Chen et~al.(2016)Chen, Zhu, Ling, Wei, and
  Jiang}]{Chen2016DistractionBasedNN}
Qian Chen, Xiaodan Zhu, Zhenhua Ling, Si~Wei, and Hui Jiang. 2016.
\newblock Distraction-based neural networks for modeling documents.
\newblock In \emph{IJCAI}.

\bibitem[{Cheng and Lapata(2016)}]{cheng-lapata:2016:P16-1}
Jianpeng Cheng and Mirella Lapata. 2016.
\newblock \href {http://www.aclweb.org/anthology/P16-1046} {Neural
  summarization by extracting sentences and words}.
\newblock In \emph{Proceedings of the 54th Annual Meeting of the Association
  for Computational Linguistics (Volume 1: Long Papers)}, pages 484--494,
  Berlin, Germany. Association for Computational Linguistics.

\bibitem[{Choi et~al.(2017)Choi, Hewlett, Uszkoreit, Polosukhin, Lacoste, and
  Berant}]{c2r:P17-1020}
Eunsol Choi, Daniel Hewlett, Jakob Uszkoreit, Illia Polosukhin, Alexandre
  Lacoste, and Jonathan Berant. 2017.
\newblock \href {https://doi.org/10.18653/v1/P17-1020} {Coarse-to-fine question
  answering for long documents}.
\newblock In \emph{Proceedings of the 55th Annual Meeting of the Association
  for Computational Linguistics (Volume 1: Long Papers)}, pages 209--220.
  Association for Computational Linguistics.

\bibitem[{Chopra et~al.(2016)Chopra, Auli, and
  Rush}]{chopra-auli-rush:2016:N16-1}
Sumit Chopra, Michael Auli, and Alexander~M. Rush. 2016.
\newblock \href {http://www.aclweb.org/anthology/N16-1012} {Abstractive
  sentence summarization with attentive recurrent neural networks}.
\newblock In \emph{Proceedings of the 2016 Conference of the North American
  Chapter of the Association for Computational Linguistics: Human Language
  Technologies}, pages 93--98, San Diego, California. Association for
  Computational Linguistics.

\bibitem[{Clarke and Lapata(2010)}]{Clarke:Lapata:2010}
James Clarke and Mirella Lapata. 2010.
\newblock Discourse constraints for document compression.
\newblock \emph{Computational Linguistics}, 36(3):411--441.

\bibitem[{Denkowski and Lavie(2014)}]{denkowski:lavie:meteor-wmt:2014}
Michael Denkowski and Alon Lavie. 2014.
\newblock Meteor universal: Language specific translation evaluation for any
  target language.
\newblock In \emph{Proceedings of the EACL 2014 Workshop on Statistical Machine
  Translation}.

\bibitem[{Efron and Tibshirani(1994)}]{efron1994introduction}
Bradley Efron and Robert~J Tibshirani. 1994.
\newblock \emph{An introduction to the bootstrap}.
\newblock CRC press.

\bibitem[{Fan et~al.(2017)Fan, Grangier, and Auli}]{control-summ}
Angela Fan, David Grangier, and Michael Auli. 2017.
\newblock \href {http://arxiv.org/abs/1711.05217} {Controllable abstractive
  summarization}.
\newblock \emph{arXiv preprint}, abs/1711.05217.

\bibitem[{Filippova et~al.(2015)Filippova, Alfonseca, Colmenares, Kaiser, and
  Vinyals}]{filippova:43852}
Katja Filippova, Enrique Alfonseca, Carlos Colmenares, Lukasz Kaiser, and Oriol
  Vinyals. 2015.
\newblock Sentence compression by deletion with lstms.
\newblock In \emph{Proceedings of the 2015 Conference on Empirical Methods in
  Natural Language Processing (EMNLP'15)}.

\bibitem[{Gillick and Favre(2009)}]{Gillick:2009:SGM:1611638.1611640}
Dan Gillick and Benoit Favre. 2009.
\newblock \href {http://dl.acm.org/citation.cfm?id=1611638.1611640} {A scalable
  global model for summarization}.
\newblock In \emph{Proceedings of the Workshop on Integer Linear Programming
  for Natural Langauge Processing}, ILP '09, pages 10--18, Stroudsburg, PA,
  USA. Association for Computational Linguistics.

\bibitem[{Gu et~al.(2017{\natexlab{a}})Gu, Cho, and
  Li}]{DBLP:journals/corr/GuCL17}
Jiatao Gu, Kyunghyun Cho, and Victor O.~K. Li. 2017{\natexlab{a}}.
\newblock \href {http://arxiv.org/abs/1702.02429} {Trainable greedy decoding
  for neural machine translation}.
\newblock In \emph{EMNLP}.

\bibitem[{Gu et~al.(2016)Gu, Lu, Li, and
  Li}]{gu-EtAl:2016:P16-1:copying_mechanism}
Jiatao Gu, Zhengdong Lu, Hang Li, and Victor~O.K. Li. 2016.
\newblock Incorporating copying mechanism in sequence-to-sequence learning.
\newblock In \emph{Proceedings of the 54th Annual Meeting of the Association
  for Computational Linguistics (Volume 1: Long Papers)}, pages 1631--1640,
  Berlin, Germany. Association for Computational Linguistics.

\bibitem[{Gu et~al.(2017{\natexlab{b}})Gu, Neubig, Cho, and
  Li}]{DBLP:journals/corr/GuNCL16}
Jiatao Gu, Graham Neubig, Kyunghyun Cho, and Victor O.~K. Li.
  2017{\natexlab{b}}.
\newblock \href {http://arxiv.org/abs/1610.00388} {Learning to translate in
  real-time with neural machine translation}.
\newblock In \emph{EACL}.

\bibitem[{Hen{\ss} et~al.(2015)Hen{\ss}, Mieskes, and
  Gurevych}]{RL_TUD-CS-2015-0145}
Sebastian Hen{\ss}, Margot Mieskes, and Iryna Gurevych. 2015.
\newblock A reinforcement learning approach for adaptive single- and
  multi-document summarization.
\newblock In \emph{International Conference of the German Society for
  Computational Linguistics and Language Technology (GSCL-2015)}, pages 3--12.

\bibitem[{Hermann et~al.(2015)Hermann, Ko\v{c}isk\'y, Grefenstette, Espeholt,
  Kay, Suleyman, and Blunsom}]{nips15_hermann}
Karl~Moritz Hermann, Tom\'a\v{s} Ko\v{c}isk\'y, Edward Grefenstette, Lasse
  Espeholt, Will Kay, Mustafa Suleyman, and Phil Blunsom. 2015.
\newblock \href {http://arxiv.org/abs/1506.03340} {Teaching machines to read
  and comprehend}.
\newblock In \emph{Advances in Neural Information Processing Systems (NIPS)}.

\bibitem[{Hirao et~al.(2013)Hirao, Yoshida, Nishino, Yasuda, and
  Nagata}]{hirao-EtAl:2013:EMNLP}
Tsutomu Hirao, Yasuhisa Yoshida, Masaaki Nishino, Norihito Yasuda, and Masaaki
  Nagata. 2013.
\newblock \href {http://www.aclweb.org/anthology/D13-1158} {Single-document
  summarization as a tree knapsack problem}.
\newblock In \emph{Proceedings of the 2013 Conference on Empirical Methods in
  Natural Language Processing}, pages 1515--1520, Seattle, Washington, USA.
  Association for Computational Linguistics.

\bibitem[{Hochreiter and Schmidhuber(1997)}]{Hochreiter:1997:LSM:265493.264179}
Sepp Hochreiter and J\"{u}rgen Schmidhuber. 1997.
\newblock Long short-term memory.
\newblock \emph{Neural Comput.}, 9(9):1735--1780.

\bibitem[{Inan et~al.(2017)Inan, Khosravi, and
  Socher}]{DBLP:journals/corr/InanKS16}
Hakan Inan, Khashayar Khosravi, and Richard Socher. 2017.
\newblock Tying word vectors and word classifiers: {A} loss framework for
  language modeling.
\newblock In \emph{ICLR}.

\bibitem[{Jing and McKeown(2000)}]{Jing:2000:CPB:974305.974329}
Hongyan Jing and Kathleen~R. McKeown. 2000.
\newblock \href {http://dl.acm.org/citation.cfm?id=974305.974329} {Cut and
  paste based text summarization}.
\newblock In \emph{Proceedings of the 1st North American Chapter of the
  Association for Computational Linguistics Conference}, NAACL 2000, pages
  178--185, Stroudsburg, PA, USA. Association for Computational Linguistics.

\bibitem[{Kikuchi et~al.(2014)Kikuchi, Hirao, Takamura, Okumura, and
  Nagata}]{kikuchi2014single}
Yuta Kikuchi, Tsutomu Hirao, Hiroya Takamura, Manabu Okumura, and Masaaki
  Nagata. 2014.
\newblock \href {http://www.aclweb.org/anthology/P14-2052} {Single document
  summarization based on nested tree structure}.
\newblock In \emph{Proceedings of the 52nd Annual Meeting of the Association
  for Computational Linguistics (Volume 2: Short Papers)}, pages 315--320,
  Baltimore, Maryland. Association for Computational Linguistics.

\bibitem[{Kim(2014)}]{kim:2014:EMNLP2014}
Yoon Kim. 2014.
\newblock Convolutional neural networks for sentence classification.
\newblock In \emph{Proceedings of the 2014 Conference on Empirical Methods in
  Natural Language Processing (EMNLP)}, pages 1746--1751, Doha, Qatar.
  Association for Computational Linguistics.

\bibitem[{Kingma and Ba(2014)}]{DBLP:journals/corr/KingmaB14}
Diederik~P. Kingma and Jimmy Ba. 2014.
\newblock \href {http://arxiv.org/abs/1412.6980} {Adam: {A} method for
  stochastic optimization}.
\newblock In \emph{ICLR}.

\bibitem[{Knight and Marcu(2000)}]{Knight:2000:SSS:647288.721086}
Kevin Knight and Daniel Marcu. 2000.
\newblock \href {http://dl.acm.org/citation.cfm?id=647288.721086}
  {Statistics-based summarization - step one: Sentence compression}.
\newblock In \emph{Proceedings of the Seventeenth National Conference on
  Artificial Intelligence and Twelfth Conference on Innovative Applications of
  Artificial Intelligence}, pages 703--710. AAAI Press.

\bibitem[{Li et~al.(2014)Li, Liu, Liu, Zhao, and Weng}]{D14-1076}
Chen Li, Yang Liu, Fei Liu, Lin Zhao, and Fuliang Weng. 2014.
\newblock \href {https://doi.org/10.3115/v1/D14-1076} {Improving
  multi-documents summarization by sentence compression based on expanded
  constituent parse trees}.
\newblock In \emph{Proceedings of the 2014 Conference on Empirical Methods in
  Natural Language Processing (EMNLP)}, pages 691--701. Association for
  Computational Linguistics.

\bibitem[{Li et~al.(2016)Li, Monroe, and Jurafsky}]{diverse}
Jiwei Li, Will Monroe, and Dan Jurafsky. 2016.
\newblock \href {http://arxiv.org/abs/1611.08562} {A simple, fast diverse
  decoding algorithm for neural generation}.
\newblock \emph{arXiv preprint}, abs/1611.08562.

\bibitem[{Lin(2004)}]{lin:2004:ACLsummarization}
Chin-Yew Lin. 2004.
\newblock \href {http://www.aclweb.org/anthology/W04-1013} {Rouge: A package
  for automatic evaluation of summaries}.
\newblock In \emph{Text Summarization Branches Out: Proceedings of the ACL-04
  Workshop}, pages 74--81, Barcelona, Spain. Association for Computational
  Linguistics.

\bibitem[{Ling and Rush(2017)}]{C2F_Summ}
Jeffrey Ling and Alexander Rush. 2017.
\newblock \href {http://aclweb.org/anthology/W17-4505} {Coarse-to-fine
  attention models for document summarization}.
\newblock In \emph{Proceedings of the Workshop on New Frontiers in
  Summarization}, pages 33--42. Association for Computational Linguistics.

\bibitem[{Louis et~al.(2010)Louis, Joshi, and
  Nenkova}]{Louis:2010:DIC:1944506.1944533}
Annie Louis, Aravind Joshi, and Ani Nenkova. 2010.
\newblock \href {http://dl.acm.org/citation.cfm?id=1944506.1944533} {Discourse
  indicators for content selection in summarization}.
\newblock In \emph{Proceedings of the 11th Annual Meeting of the Special
  Interest Group on Discourse and Dialogue}, SIGDIAL '10, pages 147--156,
  Stroudsburg, PA, USA. Association for Computational Linguistics.

\bibitem[{Luong et~al.(2015)Luong, Pham, and
  Manning}]{luong-pham-manning:2015:EMNLP}
Minh-Thang Luong, Hieu Pham, and Christopher~D. Manning. 2015.
\newblock Effective approaches to attention-based neural machine translation.
\newblock In \emph{Empirical Methods in Natural Language Processing (EMNLP)},
  pages 1412--1421, Lisbon, Portugal. Association for Computational
  Linguistics.

\bibitem[{Martins and Smith(2009)}]{Martins:2009:SJM:1611638.1611639}
Andr{\'e} F.~T. Martins and Noah~A. Smith. 2009.
\newblock \href {http://dl.acm.org/citation.cfm?id=1611638.1611639}
  {Summarization with a joint model for sentence extraction and compression}.
\newblock In \emph{Proceedings of the Workshop on Integer Linear Programming
  for Natural Langauge Processing}, ILP '09, pages 1--9, Stroudsburg, PA, USA.
  Association for Computational Linguistics.

\bibitem[{Miao and Blunsom(2016)}]{DBLP:journals/corr/MiaoB16}
Yishu Miao and Phil Blunsom. 2016.
\newblock Language as a latent variable: Discrete generative models for
  sentence compression.
\newblock In \emph{EMNLP}.

\bibitem[{Mikolov et~al.(2013)Mikolov, Sutskever, Chen, Corrado, and
  Dean}]{NIPS2013_word2vec}
Tomas Mikolov, Ilya Sutskever, Kai Chen, Greg~S Corrado, and Jeff Dean. 2013.
\newblock Distributed representations of words and phrases and their
  compositionality.
\newblock In C.~J.~C. Burges, L.~Bottou, M.~Welling, Z.~Ghahramani, and K.~Q.
  Weinberger, editors, \emph{Advances in Neural Information Processing Systems
  26}, pages 3111--3119. Curran Associates, Inc.

\bibitem[{Mnih et~al.(2016)Mnih, Badia, Mirza, Graves, Lillicrap, Harley,
  Silver, and Kavukcuoglu}]{pmlr-v48-mniha16}
Volodymyr Mnih, Adria~Puigdomenech Badia, Mehdi Mirza, Alex Graves, Timothy
  Lillicrap, Tim Harley, David Silver, and Koray Kavukcuoglu. 2016.
\newblock \href {http://proceedings.mlr.press/v48/mniha16.html} {Asynchronous
  methods for deep reinforcement learning}.
\newblock In \emph{Proceedings of The 33rd International Conference on Machine
  Learning}, volume~48 of \emph{Proceedings of Machine Learning Research},
  pages 1928--1937, New York, New York, USA. PMLR.

\bibitem[{Nallapati et~al.(2017)Nallapati, Zhai, and Zhou}]{AAAI17:summarunner}
Ramesh Nallapati, Feifei Zhai, and Bowen Zhou. 2017.
\newblock Summarunner: A recurrent neural network based sequence model for
  extractive summarization of documents.
\newblock In \emph{AAAI Conference on Artificial Intelligence}.

\bibitem[{Nallapati et~al.(2016)Nallapati, Zhou, dos santos, Gulcehre, and
  Xiang}]{nallapati2016abstractive}
Ramesh Nallapati, Bowen Zhou, Cicero~Nogueira dos santos, Caglar Gulcehre, and
  Bing Xiang. 2016.
\newblock Abstractive text summarization using sequence-to-sequence rnns and
  beyond.
\newblock In \emph{CoNLL}.

\bibitem[{Narayan et~al.(2018)Narayan, Cohen, and Lapata}]{extract_rl}
Shashi Narayan, Shay~B. Cohen, and Mirella Lapata. 2018.
\newblock \href {http://arxiv.org/abs/1802.08636} {Ranking sentences for
  extractive summarization with reinforcement learning}.
\newblock \emph{NAACL-HLT}.

\bibitem[{Noreen(1989)}]{noreen1989computer}
Eric~W Noreen. 1989.
\newblock \emph{Computer-intensive methods for testing hypotheses}.
\newblock Wiley New York.

\bibitem[{Pascanu et~al.(2013)Pascanu, Mikolov, and
  Bengio}]{pmlr-v28-pascanu13}
Razvan Pascanu, Tomas Mikolov, and Yoshua Bengio. 2013.
\newblock \href {http://proceedings.mlr.press/v28/pascanu13.html} {On the
  difficulty of training recurrent neural networks}.
\newblock In \emph{Proceedings of the 30th International Conference on Machine
  Learning}, volume~28 of \emph{Proceedings of Machine Learning Research},
  pages 1310--1318, Atlanta, Georgia, USA. PMLR.

\bibitem[{Paulus et~al.(2018)Paulus, Xiong, and
  Socher}]{DBLP:journals/corr/PaulusXS17}
Romain Paulus, Caiming Xiong, and Richard Socher. 2018.
\newblock A deep reinforced model for abstractive summarization.
\newblock In \emph{ICLR}.

\bibitem[{Press and Wolf(2017)}]{E17-2025:output_embedding}
Ofir Press and Lior Wolf. 2017.
\newblock Using the output embedding to improve language models.
\newblock In \emph{Proceedings of the 15th Conference of the European Chapter
  of the Association for Computational Linguistics: Volume 2, Short Papers},
  pages 157--163. Association for Computational Linguistics.

\bibitem[{Qian and Liu(2013)}]{qian-liu:2013:EMNLP2}
Xian Qian and Yang Liu. 2013.
\newblock \href {http://www.aclweb.org/anthology/D13-1156} {Fast joint
  compression and summarization via graph cuts}.
\newblock In \emph{Proceedings of the 2013 Conference on Empirical Methods in
  Natural Language Processing}, pages 1492--1502, Seattle, Washington, USA.
  Association for Computational Linguistics.

\bibitem[{Ranzato et~al.(2016)Ranzato, Chopra, Auli, and
  Zaremba}]{DBLP:journals/corr/RanzatoCAZ15}
Marc'Aurelio Ranzato, Sumit Chopra, Michael Auli, and Wojciech Zaremba. 2016.
\newblock Sequence level training with recurrent neural networks.
\newblock In \emph{ICLR}.

\bibitem[{Rush et~al.(2015)Rush, Chopra, and
  Weston}]{rush-chopra-weston:2015:EMNLP}
Alexander~M. Rush, Sumit Chopra, and Jason Weston. 2015.
\newblock \href {http://aclweb.org/anthology/D15-1044} {A neural attention
  model for abstractive sentence summarization}.
\newblock In \emph{Proceedings of the 2015 Conference on Empirical Methods in
  Natural Language Processing}, pages 379--389, Lisbon, Portugal. Association
  for Computational Linguistics.

\bibitem[{Schuster et~al.(1997)Schuster, Paliwal, and
  General}]{Schuster97bidirectionalrecurrent}
Mike Schuster, Kuldip~K. Paliwal, and A.~General. 1997.
\newblock Bidirectional recurrent neural networks.
\newblock \emph{IEEE Transactions on Signal Processing}.

\bibitem[{See et~al.(2017)See, Liu, and Manning}]{get_to_the_point}
Abigail See, Peter~J. Liu, and Christopher~D. Manning. 2017.
\newblock Get to the point: Summarization with pointer-generator networks.
\newblock In \emph{Proceedings of the 55th Annual Meeting of the Association
  for Computational Linguistics (Volume 1: Long Papers)}, pages 1073--1083.
  Association for Computational Linguistics.

\bibitem[{Suzuki and Nagata(2016)}]{Suzuki2016Summ}
Jun Suzuki and Masaaki Nagata. 2016.
\newblock Rnn-based encoder-decoder approach with word frequency estimation.
\newblock In \emph{EACL}.

\bibitem[{Swayamdipta et~al.(2017)Swayamdipta, Parikh, and
  Kwiatkowski}]{cascade}
Swabha Swayamdipta, Ankur~P. Parikh, and Tom Kwiatkowski. 2017.
\newblock \href {http://arxiv.org/abs/1711.00894} {Multi-mention learning for
  reading comprehension with neural cascades}.
\newblock \emph{arXiv preprint}, abs/1711.00894.

\bibitem[{Tan et~al.(2018)Tan, Wei, Yang, Lv, and Zhou}]{s-net}
Chuanqi Tan, Furu Wei, Nan Yang, Weifeng Lv, and Ming Zhou. 2018.
\newblock \href {http://arxiv.org/abs/1706.04815} {S-net: From answer
  extraction to answer generation for machine reading comprehension}.
\newblock In \emph{AAAI}.

\bibitem[{Tan et~al.(2017)Tan, Wan, and Xiao}]{graph_attn_Tan2017AbstractiveDS}
Jiwei Tan, Xiaojun Wan, and Jianguo Xiao. 2017.
\newblock Abstractive document summarization with a graph-based attentional
  neural model.
\newblock In \emph{ACL}.

\bibitem[{Vinyals et~al.(2016)Vinyals, Bengio, and Kudlur}]{OrderMatters}
Oriol Vinyals, Samy Bengio, and Manjunath Kudlur. 2016.
\newblock Order matters: Sequence to sequence for sets.
\newblock In \emph{International Conference on Learning Representations
  (ICLR)}.

\bibitem[{Vinyals et~al.(2015)Vinyals, Fortunato, and
  Jaitly}]{NIPS2015_5866:pointer_networks}
Oriol Vinyals, Meire Fortunato, and Navdeep Jaitly. 2015.
\newblock Pointer networks.
\newblock In C.~Cortes, N.~D. Lawrence, D.~D. Lee, M.~Sugiyama, and R.~Garnett,
  editors, \emph{Advances in Neural Information Processing Systems 28}, pages
  2692--2700. Curran Associates, Inc.

\bibitem[{Wang et~al.(2015)Wang, Yoshida, Hirao, Sudoh, and
  Nagata}]{Wang:2015:SBT:2876444.2876454}
Xun Wang, Yasuhisa Yoshida, Tsutomu Hirao, Katsuhito Sudoh, and Masaaki Nagata.
  2015.
\newblock \href {https://doi.org/10.1109/TASLP.2015.2432573} {Summarization
  based on task-oriented discourse parsing}.
\newblock \emph{IEEE/ACM Trans. Audio, Speech and Lang. Proc.},
  23(8):1358--1367.

\bibitem[{Williams(1992)}]{Williams:1992:SSG:139611.139614}
Ronald~J. Williams. 1992.
\newblock \href {https://doi.org/10.1007/BF00992696} {Simple statistical
  gradient-following algorithms for connectionist reinforcement learning}.
\newblock \emph{Mach. Learn.}, 8(3-4):229--256.

\bibitem[{Zajic et~al.(2004)Zajic, Dorr, and Schwartz}]{topiary}
David Zajic, Bonnie Dorr, and Richard Schwartz. 2004.
\newblock Bbn/umd at duc-2004: Topiary.
\newblock In \emph{HLT-NAACL 2004 Document Understanding Workshop}, pages
  112--119, Boston, Massachusetts.

\bibitem[{Zhou et~al.(2017)Zhou, Yang, Wei, and Zhou}]{select-enc}
Qingyu Zhou, Nan Yang, Furu Wei, and Ming Zhou. 2017.
\newblock \href {https://doi.org/10.18653/v1/P17-1101} {Selective encoding for
  abstractive sentence summarization}.
\newblock In \emph{Proceedings of the 55th Annual Meeting of the Association
  for Computational Linguistics (Volume 1: Long Papers)}, pages 1095--1104.
  Association for Computational Linguistics.

\end{thebibliography}
\bibliographystyle{acl_natbib}

\clearpage
\appendix
\section*{Supplementary Materials}
\section{Model Details}
\subsection{Convolutional Encoder}
\label{sec:conv}
Here we describe the convolutional sentence representation used in \secref{sec:sent-rep}.
We use the temporal convolutional model proposed by 
\citet{kim:2014:EMNLP2014} to compute the representation of 
every individual sentence in the document. 
First, the words are converted to the distributed vector representation by a learned word embedding matrix $W_{emb}$. 
The sequence of the word vectors from each sentence is then fed through
1-D single-layer convolution filters with various window sizes (3, 4, 5) to capture the 
temporal dependencies of nearby words and then followed by $relu$ non-linear activation and max-over-time pooling. 
The convolutional representation $r_j$ for the $j$th sentence is then obtained by concatenating the outputs from the activations of all filter window sizes.

\subsection{Abstractor}
\label{sec:copy-summ}
In this section we discuss the architecture choices for our abstractor network in \secref{sec:abs}.
At a high-level, it is a sequence-to-sequence model with attention and copy mechanism (but no coverage).
Note that the abstractor network is a separate neural network from the extractor agent without any form of parameter sharing.

\paragraph{Sequence-Attention-Sequence Model}
We use a standard encoder-aligner-decoder model~\cite{bahdanau+al-2014-nmt,luong-pham-manning:2015:EMNLP} with the bilinear multiplicative attention function~\cite{luong-pham-manning:2015:EMNLP}, 
$f_{att}(h_i, z_j) = h_i^\top W_{attn}z_j $, for the context vector $e_j$.
We share the source and target embedding matrix $W_{emb}$ as well as output projection matrix as in~\citet{DBLP:journals/corr/InanKS16,E17-2025:output_embedding,DBLP:journals/corr/PaulusXS17}.

\paragraph{Copy Mechanism}
We add the copying mechanism as in~\citet{get_to_the_point} to
extend the decoder to predict over
 the extended vocabulary of words in the input document.
A copy probability 
$p_{copy} = 
\sigma (v_{\hat{z}}^\top\hat{z}_j
+ v_s^\top z_j
+ v_w^\top w_j + b) $  
is calculated by learnable parameters $v$'s and $b$, and then is used to further compute 
a weighted sum of the probability of source vocabulary and the predefined vocabulary.
At test time, an OOV prediction is replaced by the document word with the highest 
attention score.

\subsection{Actor-Critic Policy Gradient}
\label{sec:pg}
Here we discuss the details of the actor-critic policy gradient training.
Given the MDP formulation described in \secref{sec:rl-ext}
, the return (total discounted future reward) is
\begin{equation}
\label{eq:tot-return}
R_t = \sum_{t=1}^{N_s}\gamma^{t}r(t+1)
\end{equation}
for each recurrent step $t$.
To learn a optimal policy $\pi^*$ that maximize the state-value function:
$$
V^{\pi^*}(c) =  \mathbb{E}_{\pi^*}[R_t | c_t = c]
$$
we will make use of the action-value function
$$
Q^{\pi_{\theta}}(c, j) =  \mathbb{E}_{\pi_{\theta}}
[R_t | c_t = c, j_t = j]
$$
We then take the policy gradient theorem and then substitute the action-value function with the Monte-Carlo sample:
\begin{align}
\nabla_{\theta} J(\theta) &= 
\mathbb{E}_{\pi_\theta}[
\nabla_\theta \text{log} \pi_\theta(c, j) Q^{\pi_\theta}(c, j)] \label{eq:policy_gradient_theorem}\\
&= \frac{1}{N_s} \sum_{t=1}^{N_s}\nabla_\theta \text{log} \pi_\theta(c_t, j_t) R_t
\end{align}
which runs a single episode and gets the return (estimate of action-value function) by sampling from the policy $\pi_\theta$, 
where $N_s$ is the total number of sentences the agent extracts.
This gradient update is also known as the REINFORCE algorithm \citep{Williams:1992:SSG:139611.139614}.

The vanilla REINFORCE algorithm is known for high variance. 
To mitigate this problem we add a critic network with trainable parameters $\theta_c$ having the same structure as the pointer-network's decoder (described in \secref{subsec:sent_select}) but change the final output layer to regress the state-value function $V^{\pi_{\theta_a, \omega}}(c)$.
The predicted value $b_{\theta_c, \omega}(c)$ is called the baseline and is subtracted from the action-value function to estimate the \textit{advantage}
$$ 
A^{\pi_\theta}(c, j) = Q^{\pi_{\theta_a, \omega}}(c, j) - b_{\theta_c, \omega}(c)\\
$$ 
where $\theta = \{\theta_a, \theta_c, \omega\}$
denotes the set of all trainable parameters.
The new policy gradient for our extractor can be estimated by substituting the action-value function in \eqnref{eq:policy_gradient_theorem} by the advantage and then use Monte-Carlo samples (use $R_t$ to estimate $Q$):\footnote{We found that updating with mini-batch of episodes and standardizing 
$R_t$ over all time steps and all episodes within the batch helps converging.
}
\begin{equation}
\begin{split}
\nabla_{\theta_a, \omega} J(\theta_a, \omega) \approx \\
\frac{1}{N_s} \sum_{t=1}^{N_s}
[\nabla_{\theta_a, \omega}\text{log}\pi_\theta(c, j) A^{\pi_{\theta}}(c, j)]
\end{split}
\end{equation}

\begin{figure}
\includegraphics[width=0.48\textwidth]{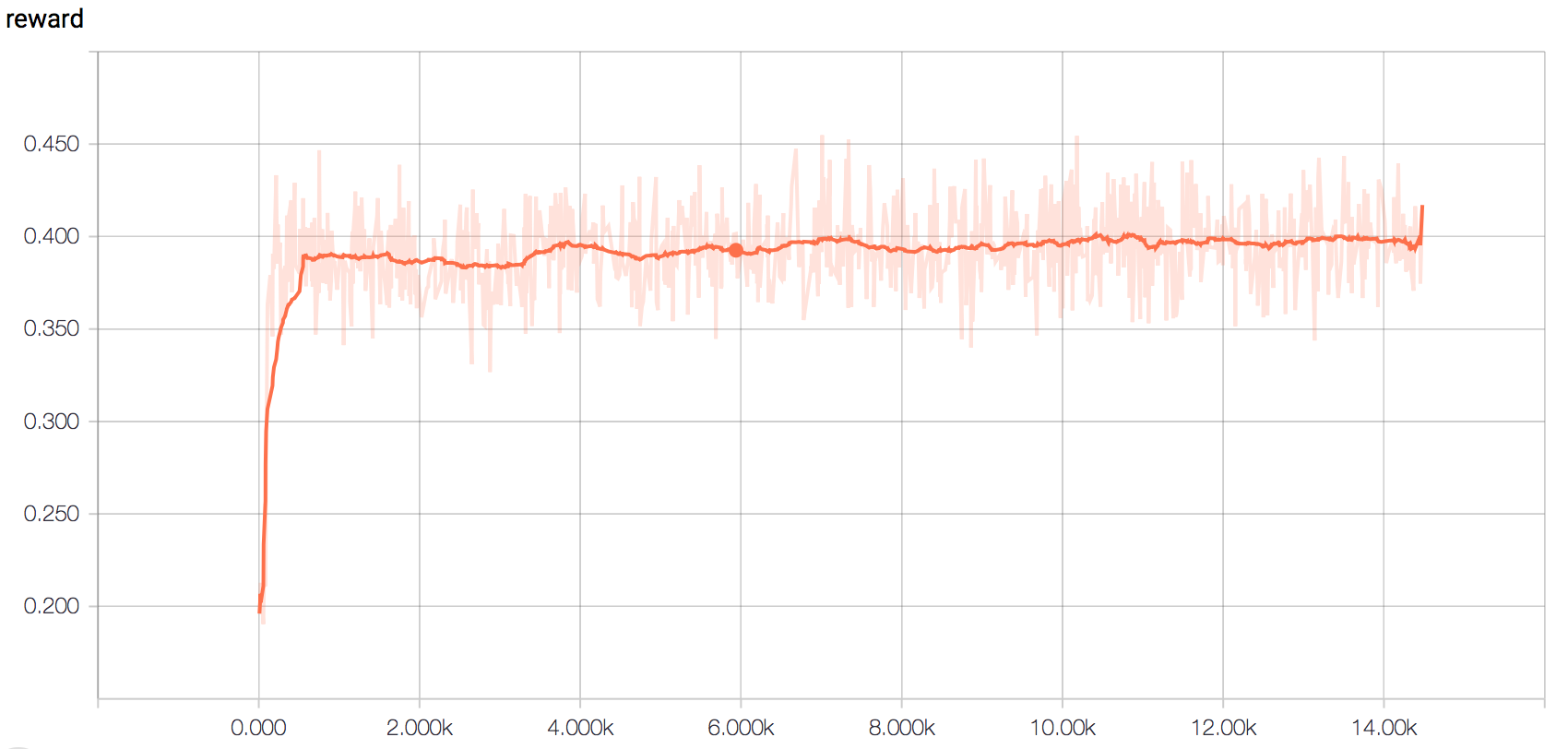}
\vspace{-20pt}
\caption{RL learning curve.}
\label{fig:rl_curve}
\vspace{-10pt}
\end{figure}

Here we also show an interesting finding of the effect adding the EOE action.
In~\figref{fig:rl_curve}, we can see that the average reward is low in the beginning but quickly goes up after the agent picks up the EOE action.
The low beginning reward is because the agent does not choose the EOE action hence keep getting zero rewards when extracting extra sentences, which lowers the average.

\subsection{Sentence Selection Baseline ff-ext}
\label{sec:ff-ext}
In this subsection, we describe the detailed network structure of the feed-forward extractor baseline (ff-ext).
Following the hierarchical sentence representation described in \secref{sec:sent-rep}, 
if we add another assumption that there exists a sequence
$j_{i1}, j_{i2}, \dots, j_{iN_s}$ where
$j_{i1} < j_{i2} < \cdots < j_{iN_s}$ such that 
\begin{align}
\label{eq:order}
 [d_{i1}, d_{i2}, \cdots, d_{iN_d}] &= x_i
 \quad \text{and} \nonumber \\
 [g(d_{j_{i1}}), g(d_{j_{i2}}), \cdots, g(d_{j_{iN_s}})] &= y_i
\end{align}
i.e., the extracted document are summarized in the order as is,
we could apply the following feed-forward structure for sentence selection.
We first learn a document representation by
\begin{equation}
\hat{x} = \tanh(W_d\frac{1}{N_d}\sum_{j=1}^{N_d}h_j + b_d)
\end{equation}
where $N_d, N_s$ each denotes the number of sentences in the document $x$ and the summary $y$ respectively. 
And then we compute the extraction probability:
$$ 
P(d_j = 1 | h_j, \hat{x})
= \sigma (W_ch_j + h_j^\top W_s\hat{x} + b)
$$
for each sentence in the document.
Assuming we have the groundtruth extraction labels 
$j_1, \dots, j_{N_s}$, 
the above formulation treats sentence selection as a sequence of 
binary classification problems, where $W$s and $b$s are trainable parameters. 
We can therefore train the sentence selection network end-to-end by cross-entropy loss, where $W$s and $b$s are trainable parameters. 

At test time, the feed-forward extractor chooses the top-k sentences and then concatenates them as the original order in the document.
Note that we still refer to this network as feed-forward extractor (ff-ext) to distinguish from the pointer network extractor (rnn-ext) though it contains recurrent structure.

\section{Training Details}

\subsection{Dataset Details}
\label{sec:data}
We use the CNN/Daily Mail dataset first proposed by \citet{nips15_hermann} 
for reading comprehension task.
This dataset has been modified for summarization by \citet{AAAI17:summarunner}.
This dataset differs from previous Gigaword dataset \cite{rush-chopra-weston:2015:EMNLP} in the length of the text: 
both documents and summaries for CNN/Daily Mail is much longer.
The standard split of the dataset contains 287,227 documents for training, 13,368 documents for validation, and 11,490 for testing.
Note that the original release of this dataset by \citet{nips15_hermann} is an anonymized version, where the
named entities are anonymized and treated as a single word in the evaluation n-gram matching.
On the other hand, \citet{get_to_the_point} proposed to use the non-anonymized, original-text version of the dataset.
For a fair comparison to prior works, we show results on both versions of the dataset.
The experiment runs training and evaluation for each version separately (but we transfer the same tuned hyperparameters from original to anonymized version).

The DUC-2002 dataset contains 567 document-summary pairs for single-document summarization. 
Due to its small size, we utilize it in a test-only setup: we directly use the CNN/Daily Mail (original text) trained model to summarize the DUC documents for testing generalization/transfer our models.
The results of \citet{get_to_the_point} on DUC is obtained by running their publicly available pretrained model.
We evaluate the results using the official ROUGE F1 script.

\subsection{Hyperparameter Details}
\label{sec:detail}
All hyper-parameters are tuned on the validation set of the original text version of CNN/DM.
We use mini-batches of 32 samples for all the training. 
Adam optimizer \citep{DBLP:journals/corr/KingmaB14}
is used with learning rate $0.001$ for ML and $0.0001$ for RL training (other hyper-parameters at their default).
We apply gradient clipping \citep{pmlr-v28-pascanu13} using 2-norm of $2.0$.
We do not use any regularization technique except early-stopping.
We also found that halving the learning rate whenever validation loss stops decreasing speeds up convergence.
For RL training, we use $\gamma = 0.95$ for the discount factor in \eqnref{eq:tot-return}.
We first train the abstractor and extractors separately until convergence with 
maximum-likelihood objectives, then apply RL training on the trained sub-modules.
For all LSTM-RNNs we use $256$ hidden units. We use single layer LSTM-RNN with $256$ hidden units for all models.
The initial states of RNN are learned for our extractor agent.
For the abstractor network, we learn a linear mapping to 
transform the encoder final states to the decoder initial states.
We also train a word2vec \citep{NIPS2013_word2vec} of 128 dimension
on the same corpus to initialize the embedding matrix for all maximum-likelihood trained models and the embedding matrix is updated during training.
We set a vocabulary size of $30000$ most common words in the training set.
For saving the memory space in training, we truncate the input article sentences to a maximum length of $100$ tokens and summary sentences to $30$ tokens (note that this is counted at the sentence-level for our abstractor training).
We use all possible sentence pairs within every summary without limit.
At test time, the length of input is not limited and the generation limit remains $30$ maximum tokens for the abstractor.
For all non-RL models, the number of sentences to extract is tuned on the validation set.
For the reranking (see \secref{sec:rerank}), we set $N = 2$ (bi-gram) and $k = 5$ (beam size).\footnote{Due to the fact that the size of the reranking list is exponential to the number of sentences of the generated summary $n$, we pruned the beam so as to allow completion (of dev-set summarization) in a reasonable amount of time, as following: for $n \leq 5$ , we use our standard beam size of $k = 5$, but for larger $n$ values, we use gradually-reduced $k$ values: $(6, 4), (7-8, 3), (9+, 2)$ for $(n, k)$.}
The diversity ratio of the diverse beam-search~\cite{diverse} is set to $1.0$.

\subsection{Training Speed}
It took a total of $19.71$ hours\footnote{4.15 hours for the abstractor, 15.56 hours for the RL training. Extractor ML training can be run at the same time with abstractor training and is approximately 1.5 hours.
} to train our model. On the other hand, \citet{get_to_the_point} reported more than 78 hours of training.
The training speed gain is mainly from the shortened input/target pairs of our abstractor model. 
Since our encoder-decoder-aligner structure operates on sentence pair, it trains much faster the the document-summary pair used in the pointer-generator model \citep{get_to_the_point}.
We also report here the speed of training our abstractor as time per training update.\footnote{We use their publicly available code and run training (without coverage mechanism) on our machine for a fair comparison.
The number of vocabulary, embedding dimension, RNN hidden units are also set to the same as our model. We set their maximum encoder and decoder steps to 400 and 100 respectively, as reported in their paper.
}
Our abstractor only requires $0.54$ seconds per updates while \citet{get_to_the_point} needs $3.42$.
For all our speed experiments we use K40 GPUs (similar to \citet{get_to_the_point}.
The reduced sequence length gives us an advantage of 6x.
Also, the model proposed by \citet{get_to_the_point} needs careful scheduling of the sentence lengths.

\begin{figure*}[t]
\centering
\begin{tabular*}{\textwidth}{| p{0.973\textwidth} | }
  \hline
  \textbf{Source document} \\ \hline
        \textsuperscript{*}[\textcolor{red}{the oxford university women 's boat race team were rescued from the thames by the royal national lifeboat institution ( rnli ) on wednesday after being overcome by choppy waters . }]
        \textsuperscript{$\mathsection$}[\textcolor{cyan}{crew members from the chiswick rnli station came to the assistance of the oxford crew and their cox , who were training for the boat race which - along with the men 's race - takes place on saturday , april 11 .}]
        \textsuperscript{$\dagger$}[\textcolor{green}{after the rowers were returned safely to putney , the sunken eight was recovered and returned to oxford 's base .}]
        \textsuperscript{$\ddagger$}[\textcolor{blue}{the royal national lifeboat institution come to the assistance of the oxford university women 's team .}]
        the oxford crew were training on the thames for the boat race which takes place on saturday , april 11 .
        the rnli revealed the conditions were caused by strong wind against the tide creating three successive waves that poured over the boat 's riggers , ` creating an influx of water that could not be managed by the craft 's bilge pump ' .
        in a statement rnli helmsman ian owen said : ` while we have rescued quite a number of rowers over the years , this is the first time i 've been involved in helping such a prestigious team .
        ` the weather can be unpredictable on the thames , and the oxford university team dealt with the situation as safely and calmly as possible . we wish them all the best for their upcoming race . '
        chiswick and tower stations are the busiest in the country , and the rnli has saved over 3,600 people since the service began in 2002 .
        the rnli alternative boat race fundraising event on april 10 takes place the day before the bny mellon boat race on the same famous stretch of river . for more information , please visit : rnli.org / boatrace .

  \\ \hline
  \textbf{Ground truth summary} \\ \hline
the crew were training for the boat race which takes place on april 11 . \\
the sunken eight was recovered and returned to oxford 's base . \\
the choppy conditions were caused by strong wind against the tide creating three successive waves that poured over the boat 's riggers . \\
  \hline
  \textbf{rnn-ext + abs + RL} (ROUGE-1: 48.54, ROUGE-2: 27.72 ROUGE-L: 48.54)\\ \hline
  \textsuperscript{*}\textcolor{red}{
the oxford university women 's boat race team were rescued from the thames by the royal national lifeboat institution .} \\
\textsuperscript{$\mathsection$}\textcolor{cyan}{
crew members were training for the boat race which takes place on saturday .
} \\
\textsuperscript{$\dagger$}\textcolor{green}{
the rowers were returned to oxford 's base .
} \\
\textsuperscript{$\ddagger$}\textcolor{blue}{
the royal national lifeboat institution come to the assistance of the oxford university women 's team .
} \\ \hline
  \textbf{+rerank} (ROUGE-1: 60.42, ROUGE-2: 42.55, ROUGE-L: 60.42)\\ \hline
  \textsuperscript{*}\textcolor{red}{
the oxford university women 's boat race team were rescued from the thames .
  } \\
\textsuperscript{$\mathsection$}\textcolor{cyan}{
crew members were training for the boat race which takes place on saturday .
} \\
\textsuperscript{$\dagger$}\textcolor{green}{
the sunken eight was recovered and returned to oxford 's base .
} \\
\textsuperscript{$\ddagger$}\textcolor{blue}{
the royal national lifeboat institution come to the assistance of the team .
} \\
  \hline
\end{tabular*}
\caption{
Example from the dataset showing the generated summary of our best models. 
The colored (marked) sentences correspond to our extractor's sentence selection.
The listed ROUGE scores are computed for this specific example.
}
\label{fig:sample1}
\end{figure*}

\begin{figure*}[t]
\centering
\begin{tabular*}{\textwidth}{| p{0.973\textwidth} | }
  \hline
  \textbf{Source document} \\ \hline
        ( cnn ) have mercy ! lifetime has its follow-up to its `` unauthorized saved by the bell '' tv movie : the network is now taking on full house .
        \textsuperscript{*}[\textcolor{red}{the female-skewing cable network has greenlit `` the unauthorized full house story '' ( working title ) , the hollywood reporter has learned .}]
        \textsuperscript{$\mathsection$}[\textcolor{cyan}{in the same vein as its `` saved by the bell '' pic , lifetime 's full house story will look at the rise of the cast -- including john stamos , bob saget and the mary-kate and ashley olsen -- and explore the pressure they faced to balance idyllic family life on the show with the more complicated reality of their own lives outside the series . additionally , it will look at the warm bond that grew between the cast as the show became one of america 's most beloved family sitcoms .}]
        \textsuperscript{$\dagger$}[\textcolor{blue}{casting will begin immediately . an air date for the `` full house '' tell-all has yet to be determined .}]
        see more broadcast tv 's returning shows 2015-16 .
        \textsuperscript{$\ddagger$}[\textcolor{green}{ron mcgee , who penned the `` unauthorized saved by the bell story , '' will write the `` full house '' take . the telepic will be produced by the bell team of front street pictures and ringaling productions , with harvey kahn and stephen bulka also on board to exec produce .}]
        for lifetime , the news comes after its two-hour bell take fizzled on labor day 2014 . despite tons of build-up and excitement from diehard fans of the original comedy series , the bell take drew only 1.6 million total viewers , with 1.1 million viewers among the 18-49 and 25-54 demographics . that pic was based on former star dustin diamond 's behind the bell 2009 tell-all , with dylan everett starring as mark-paul gosselaar and sam kindseth as diamond .
        full house aired on abc from 1987 to 1995 . netflix this month revived the beloved family comedy as `` fuller house , '' with original stars candace cameron-bure ( d.j. ) , her on-screen sister , jodie sweetin ( stephanie ) , and best friend andrea barber ( kimmy ) , in a 13-episode follow-up series .
        from its start as an unassuming family comedy in 1987 to its eventual wildly popular 192-episode run , `` full house '' was `` the little sitcom that could . '' it made huge stars of its cast -- from bob saget and dave coulier , who were grinding away on the standup circuit , to john stamos breaking hearts on general hospital , and the olsen twins .
        see the original story at the hollywood reporter 's website .
        2015 the hollywood reporter . all rights reserved .

  \\ \hline
  \textbf{Ground truth summary} \\ \hline
          the network has reportedly greenlit the tell-all . \\
        lifetime previously did an unauthorized movie on `` saved by the bell '' \\
  \hline
  \textbf{rnn-ext + abs + RL} (ROUGE-1: 25.00, ROUGE-2: 7.41 ROUGE-L: 25.00)\\ \hline
  \textsuperscript{*}\textcolor{red}{the female-skewing cable network has greenlit `` the unauthorized full house story ''} \\
\textsuperscript{$\mathsection$}\textcolor{cyan}{the cast will look at the warm bond that grew between the cast .} \\
\textsuperscript{$\ddagger$}\textcolor{green}{ron mcgee will write the `` full house '' take .} \\
\textsuperscript{$\dagger$}\textcolor{blue}{casting will begin immediately .} \\ \hline
  \textbf{+rerank} (ROUGE-1: 37.93, ROUGE-2: 17.86, ROUGE-L: 37.93)\\ \hline
  \textsuperscript{*}\textcolor{red}{the female-skewing cable network has greenlit `` the unauthorized full house story ''} \\
\textsuperscript{$\mathsection$}\textcolor{cyan}{lifetime 's full house story will look at the rise of the cast .} \\
\textsuperscript{$\ddagger$}\textcolor{green}{ron mcgee penned the `` unauthorized saved by the bell story ''} \\
\textsuperscript{$\dagger$}\textcolor{blue}{casting will begin immediately .} \\
  \hline
\end{tabular*}
\caption{
Example from the dataset showing the generated summary of our best models. 
The colored (marked) sentences correspond to our extractor's sentence selection.
The listed ROUGE scores are computed for this specific example.
}
\label{fig:sample2}
\end{figure*}

\section{Generation Samples}
\label{sec:samples}
Please see~\figref{fig:sample1}  and ~\figref{fig:sample2} for the output examples (see the discussion of this example in \secref{sec:qualitative}).
\clearpage

\end{document}